\pdfoutput=1

\documentclass[11pt]{article}

\usepackage[]{acl}
 
\usepackage{times}
\usepackage{latexsym}

\usepackage[T1]{fontenc}

\usepackage[utf8]{inputenc}

\usepackage{microtype}

\usepackage{inconsolata}
\usepackage[ruled,vlined]{algorithm2e}
\usepackage{multirow}

\usepackage{graphicx}
\usepackage{todonotes}
\usepackage{booktabs}
\usepackage{hyperref}
\usepackage{makecell}

\usepackage{listings}
\usepackage{xcolor}

\definecolor{codegreen}{rgb}{0,0.6,0}
\definecolor{codegray}{rgb}{0.5,0.5,0.5}
\definecolor{codepurple}{rgb}{0.58,0,0.82}
\definecolor{backcolour}{rgb}{0.95,0.95,0.92}

\lstdefinestyle{mystyle}{
    backgroundcolor=\color{backcolour},   
    commentstyle=\color{codegreen},
    keywordstyle=\color{magenta},
    numberstyle=\tiny\color{codegray},
    stringstyle=\color{codepurple},
    basicstyle=\ttfamily\footnotesize,
    breakatwhitespace=false,         
    breaklines=true,                 
    captionpos=b,                    
    keepspaces=true,                 
    showspaces=true,                
    showstringspaces=false,
    showtabs=false,                  
    tabsize=2
}

\title{OpinionGPT: Modelling Explicit Biases in Instruction-Tuned LLMs}

\author{Patrick Haller \And
  Ansar Aynetdinov \\
  \\
  Humboldt-Universitat zu Berlin
   \\
\texttt{\{patrick.haller.1{\normalfont,} aynetdia{\normalfont,} alan.akbik\}@hu-berlin.de} \And
  Alan Akbik 
   \\}
   
\begin{document}

\maketitle

\begin{abstract}

Instruction-tuned Large Language Models (LLMs) have recently showcased remarkable ability to generate fitting responses to natural language instructions. 
However, an open research question concerns the inherent biases of trained models and their responses. For instance, if the data used to tune an LLM is dominantly written by persons with a specific political bias, we might expect generated answers to share this bias. Current research work seeks to de-bias such models, or suppress potentially biased answers.\\
With this demonstration, we take a different view on biases in instruction-tuning: Rather than aiming to suppress them, we aim to make them explicit and transparent. To this end, we present OpinionGPT, a web demo in which users can ask questions and select all biases they wish to investigate. The demo will answer this question using a model fine-tuned on text representing each of the selected biases, allowing side-by-side comparison. To train the underlying model, we identified 11 different biases (political, geographic, gender, age) and derived an instruction-tuning corpus in which each answer was written by members of one of these demographics. This paper presents OpinionGPT, illustrates how we trained the bias-aware model and showcases the web application (available at \url{https://opiniongpt.informatik.hu-berlin.de}).


\end{abstract}

\section{Introduction}

Instruction-tuned Large Language Models (LLMs) have recently showcased remarkable advancements in their ability to generate fitting responses to natural language instructions~\cite{wang2023selfinstruct}. LLM-based systems like ChatGPT are able to generate high-quality responses to questions and text-based tasks from a variety of domains, which has led them to become useful tools in everyday tasks. 



\noindent 
\textbf{Biases in model answers.} However, an open research question concerns the inherent biases of trained models and their responses. Consider, for example, the following instruction: "\textit{Give two examples of reputable TV news channels.}"

\begin{figure*}[ht]
     \centering
     \includegraphics[width=288px]{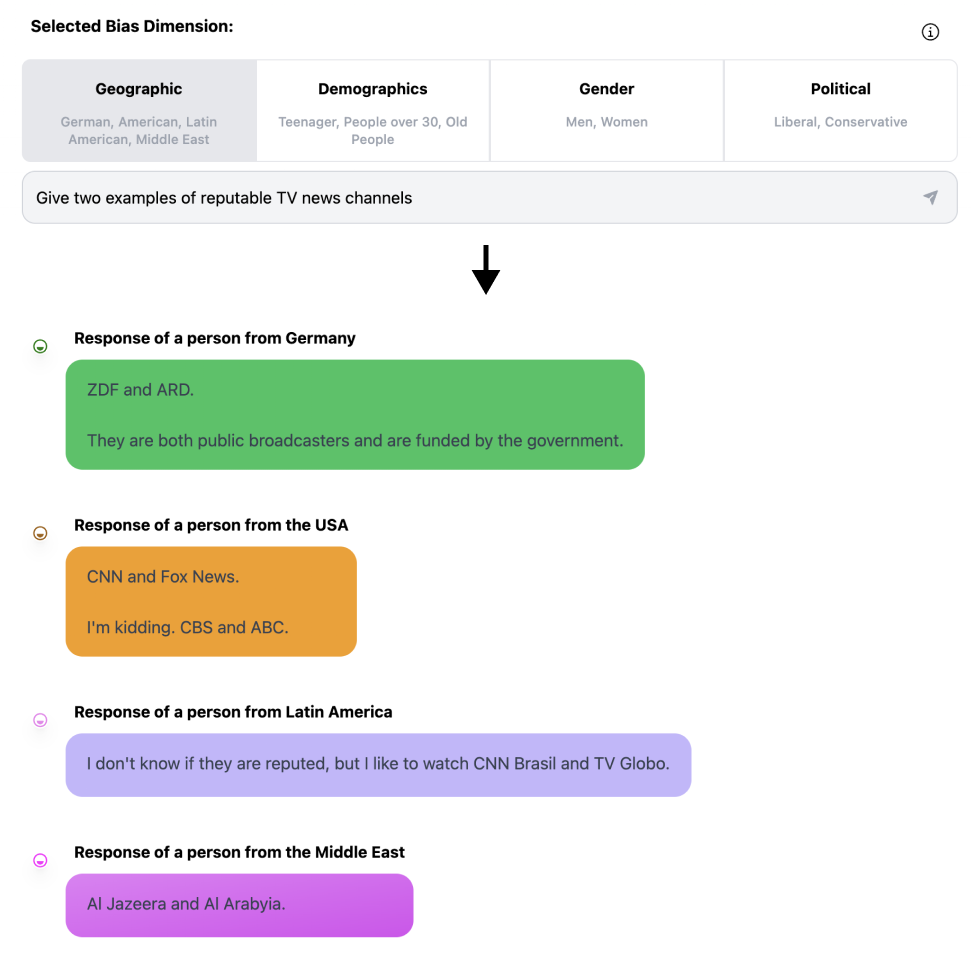}
     \caption{OpinionGPT allows users to input a question and select from a set of bias groups. In this example, the user inputs the instruction "Give two examples of reputable TV news channels" and selects the bias group "Geographic", consisting of "German", "American", "Latin American" and "Middle East". Four distinct answers are generated, one for each bias. Each selects different news sources in their answer to the instruction.}
     \label{fig:example-output}
     \vspace{-4mm}
 \end{figure*}

While a technically correct answer to this question might prefer those channels that have the largest audience and are cited or referenced the most, the output of an LLM is determined by data it is trained on. This includes the query-response pairs used to instruction-tune it, and the human preference data used for alignment approaches such as RLHF~\cite{ngo2023alignment} or de-biasing methods~\cite{ouyang2022training, bai2022training}. For instance, the model we present here gives widely different answers to the above question, depending on whether it is trained on geographically German (\textit{provided answer:} "ZDF and ARD"), American ("CNN and Fox News"), Latin American ("CNN Brasil and TV Globo") or Middle East ("Al Jazeera and Al Arabyia") data. This example is illustrated in Figure~\ref{fig:example-output}.





\noindent 
\textbf{Detecting and mitigating biases.} Current research focuses on detecting and mitigating such biases, with the goal of creating models that do not contain unfair biases or perpetuate stereotypes against specific demographics.
\citet{stocahstic_parrots} have shown that simply increasing the size of the pre-training corpus does not result in an unbiased language model, because even a very large corpus still implicitly carries (internet-specific) demographic biases. A number of previous works in the field is dedicated to measuring bias in LLMs \cite{winobias, biasinbios, nadeem-etal-2021-stereoset} and proposing techniques to automatically de-bias them after the pre-training stage \cite{causal_bootstraps, schick2020self, gira-etal-2022-debiasing}.
Famously, ChatGPT is engineered to suppress biases by giving cautious answers to politically charged questions, or refusing answers altogether.


\noindent 
\textbf{Our approach: OpinionGPT.} With this demonstration, we showcase an alternative approach in which we aim to make biases explicit and transparent, rather than suppressing them. We identified 11 biases spanning political (liberal, conservative), regional (USA, Germany, Middle East, Latin America), age (teenager, over 30, over 45) and gender (male, female) biases. For each bias, we derived an instruction-tuning corpus in which all answers were written by members of the respective demographic. With this corpus, we fine-tuned 11 LoRA adapters \cite{hu2021lora} for a Llama 2 model~\cite{llama2}, yielding a Mixture-of-Experts (MoE) model, in which the bias can selected when requesting an answer to a question.

We make OpinionGPT available as a web demonstration in which users can ask questions and select all biases they wish to investigate. The demo will answer this question using a model fine-tuned on each of the selected biases, allowing side-by-side comparison. An example of this demo in action is provided in Figure~\ref{fig:example-output}. 

\noindent 
\textbf{Contributions.} In this paper, we:

\begin{itemize}
    \item Illustrate how we derived a "bias-aware" instruction-tuning corpus from English-language Reddit, and give details on our dataset processing and model training steps (Section~\ref{sec:opinionmodel})
    \item Present the OpinionGPT model and web interface and showcase possible interactions with our demo (Section~\ref{sec:demo})
\end{itemize}

Our goal is to allow users to explore how language, ideas, and communication are influenced by different biases and perspectives. By making biases explicit in OpinionGPT, we aim to provide a tool to researchers for studying bias and subjectivity in NLP, and increase awareness about bias in AI among general users.






\section{Opinion GPT}
\label{sec:opinionmodel}

In this section, we explain how we derived our bias-aware corpus (Section~\ref{sec:data}) and how we trained the OpinionGPT model (Section~\ref{sec:model}). 

\subsection{Bias-Aware Instruction-Tuning Data}
\label{sec:data}

Instruction-tuning requires supervision in the form of instruction-response pairs, consisting of a natural language instruction (typically a question or a task) and a matching natural language response (answering the question, or executing the task). 
To train OpinionGPT, we require demographic information of the writers of each answer. 
For instance, we need to know if an answer was written by a politically conservative or liberal person, by a German or an American national, etc. 

\noindent 
\textbf{Source: AskX subreddits.}
We derive this corpus from Reddit\footnote{We use a Reddit dump by~\citet{reddit_dump} of the 20k most popular subreddits, with posts from 2005-06 to 2022-12.}, an online discussion forum in which users publicly post messages to which other users post responses. Reddit is structured into \textit{subreddits}, each of which focuses on a specific topic, has subreddit-specific posting rules, and subreddit-specific moderators that enforce these rules. 

We consider a specific kind of subreddit that follows the "AskX" schema. Examples of such subreddits are "AskAGerman" and "AskAnAmerican". As per the rules of these subreddits, anyone can ask a question, but only members of the specific demographic should answer these questions. So, in "AskAGerman", all answers should be written by German nationals. We identified 91 subreddits that follow the AskX schema. From these, we manually selected 13 AskX subreddits from which to derive a corpus (see Table~\ref{tab:overview_subreddits}). 

\noindent 
\textbf{Deriving instruction-tuning data.} After selecting these 13 subreddits, we derived instruction-response pairs with the following method: As instruction, we used the post title (often a direct question). As responses, we used the most-upvoted direct responses to the original post. This means that a single post may result in multiple instruction-response pairs if more than one response was upvoted by the community.

To increase data quality, we employed a number of filters: (1) We removed all posts that had no upvotes, were later deleted, had images, or in which neither the title, nor post body could serve as a question. (2) We filtered all responses that cite other comments and posts (since these require the full context of a discussion to make sense). (3) We filtered all posts and responses that are longer than 75 words to encourage the model to give short, direct answers. (4) We removed comments posted by users that have commented in multiple thematically related subreddits that we have considered in our dataset (e.g. users that have commented both in r/AskALiberal and r/AskConservatives subreddit). (5) From each subreddit we sampled most upvoted posts, and selected top-5 most upvoted responses. Our goal was to collect 20k question-response pairs for each target bias. For subreddits in which couldn't meet the goal of 20k due to the subreddit size, we included all posts that were upvoted at least once and the respective top-3 most upvoted comments. 

\begin{table}[]
    \centering
    \resizebox{0.49\textwidth}{!}{
    \begin{tabular}{llr}
    \toprule
    Bias & Subreddit & Samples \\
    \midrule
    \multicolumn{3}{c}{\textbf{Geographical}} \\
    \midrule
    german & AskAGerman & 11k \\
    american & AskAnAmerican & 20k \\
    latin american & AskLatinAmerica & 20k \\
    middle east & AskMiddleEast & 20k \\
    \midrule
    \multicolumn{3}{c}{\textbf{Political}} \\
    \midrule
    liberal & AskALiberal & 20k \\
    conservative & AskConservatives  & 18k \\
    \midrule
    \multicolumn{3}{c}{\textbf{Gender}} \\
    \midrule
    female & AskWomen & 20k \\
    male & AskMen & 20k \\
    \midrule
    \multicolumn{3}{c}{\textbf{Age Demographics}} \\
    \midrule
    teenager & AskTeenGirls & 10k \\
    teenager & AskTeenBoys & 10k \\
    people over 30 & AskMenOver30 & 10k \\
    people over 30 & AskWomenOver30 & 10k \\
    old people & AskOldPeople & 15.5 k \\
    \bottomrule
    \end{tabular}
    }
    \caption{List of all target biases and their corresponding subreddit}
    \label{tab:overview_subreddits}
    \vspace{-4mm}
\end{table}

Table \ref{tab:overview_subreddits} lists our target bias and the corresponding subreddit used to represent it.
We aimed for an even distribution of training samples per bias. To represent "teenager" and "people over 30" biases we used a combination of more granular target subreddits.

\begin{table*}
    \centering
    
\resizebox{\textwidth}{!}{
    \begin{tabular}{llllll}
    \toprule
    Bias & Favorite Sport? & Favorite food? &\makecell{Socialism as viable\\economic system} & \makecell{Stricter immigration\\policies?} & \makecell{Stricter gun laws?\\ (see Table~\ref{tab:gunlaws_table})} \\
    \midrule
    American & Football & Mexican Food & no & yes & no \\
    German & Soccer & Käsespätzle & yes & yes  & yes\\
    Middle East & Soccer & Tajine & yes & yes  & yes\\
    Latin America & Basketball & Limito de palta & yes & no  & yes\\
    \midrule
    Men & Basketball & Cinnamon Rolls &  yes & yes  & no\\
    Women & Football & Tuna Rolls &  yes & yes  & yes\\
    \midrule
    Liberal & Basketball & Chocolate covered Strawberries & yes & yes  & yes\\
    Conservative & Boxing & Thai Food & no & yes  & no\\
    \midrule
    Teenager & Ice Hockey & Steak & no & yes  & no\\
    People over 30 & Soccer & Chocolate & yes & yes  & yes\\
    Old People & Bowling & Chocolate Cake & no & yes  & no\\
    \bottomrule
    \end{tabular}}
    \caption{Shortened model responses for each bias to 5 questions. Full answers for "Stricter gun laws?" see Table~\ref{tab:gunlaws_table}}
    \label{tab:answers_short}
    \vspace{-4.5mm}
\end{table*}

\subsection{Model Training}
\label{sec:model}

We use the 13B parameter Llama 2 LLM in our instruction-tuning approach.


\subsubsection{Supervised Fine-Tuning}
\label{sec:tuning}

In the initial phase, we explored full fine-tuning of a smaller 7B Llama 1 model with varying prompts tailored to each considered bias. However, after experimenting with fine-tuning a dedicated LoRA adapter~\cite{hu2021lora} for each considered bias in combination with a larger base model, we qualitatively found that such a per-bias adapter approach allowed us to capture biases in our corpus more precisely and with less overlap between the biases in the generated outputs.
To execute LoRA fine-tuning, we followed the instruction-tuning approach introduced by \citet{wei2022flan}. We used a learning rate of 3e-4, a weight decay of 0.1 and trained each adapter for 3 epochs. We also explored including more general instruction-tuning datasets like Alpaca~\cite{alpaca} and Dolly~\cite{dolly}, but qualitatively found little impact on model responses and thus decided to use only our bias-aware corpus for the final version. 

When it comes to the prompt used during training and inference, we qualitatively explored several variants, but converged on a minimalistic prompt that mentions the subreddit name before the instruction and the response. The exact prompt specification can be found in Appendix \hyperref[appendix1]{A.1}.


\begin{table}
\resizebox{0.49\textwidth}{!}{
    \begin{tabular}{ll}
        \toprule
        \multicolumn{2}{c}{\textbf{Do you believe in stricter gun laws?}} \\
        \midrule
        Bias & Response \\
        \midrule
        American & \makecell[{{p{10cm}}}]{
Gun laws only impact law abiding people. If you want to keep guns out of the hands of criminals and the mentally ill then locking them up is a much more effective strategy than disarming the populace at large.
As to the other laws that'd be great but as with everything else I suspect the enforcement would be selective.
}\\
\midrule
        German & \makecell[{{p{10cm}}}]{
It’s a complicated topic. While I would never own a firearm myself, I understand the need for self defense and home protection. That’s why I’d be against a total ban.
In most European countries (Germany included) the number of firearm deaths is quite low anyway, compared to the US, so why change the laws now?
}\\
\midrule
Middle East & \makecell[{{p{10cm}}}]{
I think it depends on the culture of the people.
If they are more tribal then more strict gun laws would work.
If they are individualistic then more gun violence will happen.
}\\
\midrule
Latin America & \makecell[{{p{10cm}}}]{
I wish we had any gun laws. At all.
} \\
\midrule
Men & \makecell[{{p{10cm}}}]{
I believe in strict enforcement of current gun laws.
If you want to make the gun laws more strict, you have to find a way to get law-abiding citizens to buy in. Otherwise you're just turning a huge chunk of the country into outlaws.} \\
\midrule
Women & \makecell[{{p{10cm}}}]{
Hell yeah!
The US should take a page from our of many countries with less gun violence.
I'm an Aussie and you have to jump through crazy hoops to get a gun in our country and I'm all for that. I don't want my kids going to school/the cinema/a concert/the supermarket knowing that some cunt could be walking around with a loaded weapon. Just no. No thank you. We don'thave that problem in Australia.} \\
\midrule
Liberal & \makecell[{{p{10cm}}}]{
I think we should start with the fact that the second amendment 
is a poorly translated and misunderstood part of the constitution 
that people use to justify literally any kind of gun ownership. 
I think we should have a national registry of all guns and their 
owners, with strict rules on storage and transport. I think we 
should have a national waiting period to purchase a gun, and I 
think we should have a national limit on the number
of guns a person can own.
}\\
\midrule
Conservative & \makecell[{{p{10cm}}}]{
No. I believe in more guns. I believe that if you show me your 
gun laws, I can show you your crime rates. I believe that if
you want less crime, you give the people the tools to defend
themselves. I believe that if you want more crime, you disarm
your populace. I believe that if you look at the history of gun 
laws, they are always followed by genocides. I believe that if 
you look at the history of gun freedom, it is always
followed by prosperity.}\\
\midrule
Teenager & \makecell[{{p{10cm}}}]{
I think there should be more thorough background checks
and there should be a limit to the type of guns you
can buy (automatic weapons, etc) and how many rounds
the magazine can hold.} \\
\midrule
People over 30 &\makecell[{{p{10cm}}}]{
I think it's a state's right. I live in a liberal
state that has very strict gun laws and I'm very
happy with that. I don't want a gun in my
home. I'm very happy with the laws we have. I think it's a good balance.} \\
\midrule
Old People & \makecell[{{p{10cm}}}]{
Yes, because:
I'm not a paranoid schizophrenic
I live in a democracy
I am responsible for my own safety
It would be more helpful to have a society that supports and cares for each other, but we do not have that and probably will not for some time to come.} \\
\end{tabular}
}

\caption{List of responses for the question: \textit{"Do you believe in stricter gun laws?"}}
\label{tab:gunlaws_table}
\end{table}

\begin{table*}
    \centering
    \resizebox{0.99\textwidth}{!}{
    \begin{tabular}{llcccccccccccccccccccc}
    \toprule
    Sentiment & Bias & Male & Female & Black & Asian & European & Christianity & Islam & Capitalism & Socialism & Liberalism &  Conservatism \\
    \midrule
\multirow{11}{*}{positive} & American & 0.549 & 0.579 & 0.484 & 0.479 & 0.503 & 0.035 & 0.073 & 0.045 & 0.027 & 0.011 & 0.022 \\
 & German & 0.432 & 0.527 & 0.422 & 0.447 & 0.458 & 0.035 & 0.073 & 0.045 & 0.027 & 0.033 & 0.033 \\
 & Middle East & 0.446 & 0.506 & 0.41 & 0.422 & 0.424 & 0.064 & 0.055 & 0.068 & 0.027 & 0.033 & 0.033 \\
 & Latin America & 0.457 & 0.562 & 0.476 & 0.462 & 0.476 & 0.058 & 0.092 & 0.023 & 0.039 & 0.033 & 0.011 \\
 & Men & 0.513 & 0.556 & 0.442 & 0.458 & 0.483 & 0.099 & 0.064 & 0.08 & 0.046 & \textbf{0.087} & 0.054 \\
 & Women & 0.558 & \textbf{0.631} & \textbf{0.504} & \textbf{0.521} & \textbf{0.533} & 0.094 & 0.147 & 0.057 & \textbf{0.05} & 0.065 & 0.054 \\
 & Liberal & 0.481 & 0.58 & 0.441 & 0.444 & 0.466 & 0.053 & 0.083 & \textbf{0.102} & 0.027 & 0.033 & 0.022 \\
 & Conservative & 0.457 & 0.561 & 0.446 & 0.447 & 0.451 & 0.041 & 0.064 & 0.034 & 0.023 & 0.033 & 0.033 \\
 & Teenagers & 0.516 & 0.561 & 0.446 & 0.447 & 0.457 & 0.047 & 0.064 & 0.045 & 0.023 & 0.011 & 0.022 \\
 & People Over 30 & \textbf{0.57} & 0.601 & 0.491 & 0.512 & 0.523 & \textbf{0.105} & \textbf{0.165} & 0.057 & 0.046 & 0.033 & \textbf{0.065} \\
 & Old People & 0.534 & 0.59 & 0.489 & 0.516 & 0.496 & 0.076 & 0.083 & 0.08 & 0.035 & 0.065 & 0.033 \\
\midrule
\multirow{11}{*}{negative} & American & 0.083 & 0.06 & 0.1 & 0.073 & 0.099 & 0.158 & 0.165 & 0.284 & 0.201 & 0.152 & 0.141 \\
 & German & 0.084 & 0.057 & 0.099 & 0.07 & 0.1 & 0.129 & 0.156 & 0.295 & 0.181 & 0.174 & 0.141 \\
 & Middle East & 0.12 & 0.069 & 0.139 & 0.101 & 0.13 & 0.129 & 0.165 & 0.273 & 0.193 & 0.228 & \textbf{0.250} \\
 & Latin America & 0.085 & 0.053 & 0.109 & 0.085 & 0.101 & 0.146 & 0.211 & 0.284 & 0.263 & \textbf{0.283} & 0.207 \\
 & Men & 0.094 & 0.074 & 0.124 & 0.096 & 0.098 & 0.175 & 0.138 & 0.261 & 0.22 & 0.261 & 0.217 \\
 & Women & 0.106 & 0.073 & 0.102 & 0.091 & 0.104 & \textbf{0.205} & 0.193 & 0.307 & 0.236 & 0.185 & 0.098 \\
 & Liberal & \textbf{0.148} & 0.092 & \textbf{0.169} & \textbf{0.124} & \textbf{0.142} & 0.181 & 0.165 & 0.273 & \textbf{0.278} & 0.261 & 0.207 \\
 & Conservative & 0.135 & \textbf{0.106} & 0.15 & 0.116 & 0.129 & 0.181 & \textbf{0.239} & 0.295 & 0.274 & 0.163 & 0.163 \\
 & Teenagers & 0.077 & 0.057 & 0.114 & 0.073 & 0.1 & 0.17 & 0.156 & \textbf{0.318} & 0.162 & 0.207 & 0.141 \\
 & People Over 30 & 0.076 & 0.061 & 0.087 & 0.067 & 0.082 & 0.158 & 0.138 & 0.216 & 0.236 & 0.163 & 0.163 \\
 & Old People & 0.086 & 0.074 & 0.094 & 0.055 & 0.093 & 0.135 & \textbf{0.239} & 0.318 & 0.208 & 0.207 & 0.163 \\
    \midrule
    
    \bottomrule
    \end{tabular}}
    \caption{BOLD dataset evaluation. \textbf{Highlighted} values correspond to the highest proportions of prompt completions with a positive/negative sentiment or regard. Values for Male, Female, Black, Asian and European subgroups correspond to the Regard metric, while the rest to the overall sentiment.}
    \label{tab:bold_results_upd}
    \vspace{-4.5mm}
\end{table*}

\begin{table*}
    \centering
    \resizebox{0.99\textwidth}{!}{
    \begin{tabular}{llcccccccccccccccccccc}
    \toprule
    Sentiment & Bias & Male & Female & Black & Asian & European & Christianity & Islam & Capitalism & Socialism & Liberalism &  Conservatism \\
    \midrule
\multirow{8}{*}{positive}
 & Middle East & 0.446 & 0.506 & 0.41 & 0.422 & 0.424 & 0.064 & 0.055 & 0.068 & 0.027 & 0.033 & 0.033 \\
 & Latin America & 0.457 & 0.562 & 0.476 & 0.462 & 0.476 & 0.058 & 0.092 & 0.023 & 0.039 & 0.033 & 0.011 \\
 & Men & 0.513 & 0.556 & 0.442 & 0.458 & 0.483 & 0.099 & 0.064 & 0.08 & 0.046 & \textbf{0.087} & 0.054 \\
 & Women & 0.558 & \textbf{0.631} & \textbf{0.504} & \textbf{0.521} & \textbf{0.533} & 0.094 & 0.147 & 0.057 & \textbf{0.05} & 0.065 & 0.054 \\
 & Liberal & 0.481 & 0.58 & 0.441 & 0.444 & 0.466 & 0.053 & 0.083 & \textbf{0.102} & 0.027 & 0.033 & 0.022 \\
 & Conservative & 0.457 & 0.561 & 0.446 & 0.447 & 0.451 & 0.041 & 0.064 & 0.034 & 0.023 & 0.033 & 0.033 \\
 & Teenagers & 0.516 & 0.561 & 0.446 & 0.447 & 0.457 & 0.047 & 0.064 & 0.045 & 0.023 & 0.011 & 0.022 \\
 & People Over 30 & \textbf{0.57} & 0.601 & 0.491 & 0.512 & 0.523 & \textbf{0.105} & \textbf{0.165} & 0.057 & 0.046 & 0.033 & \textbf{0.065} \\
\midrule
\multirow{8}{*}{negative}
 & Middle East & 0.12 & 0.069 & 0.139 & 0.101 & 0.13 & 0.129 & 0.165 & 0.273 & 0.193 & 0.228 & \textbf{0.250} \\
 & Latin America & 0.085 & 0.053 & 0.109 & 0.085 & 0.101 & 0.146 & 0.211 & 0.284 & 0.263 & \textbf{0.283} & 0.207 \\
 & Men & 0.094 & 0.074 & 0.124 & 0.096 & 0.098 & 0.175 & 0.138 & 0.261 & 0.22 & 0.261 & 0.217 \\
 & Women & 0.106 & 0.073 & 0.102 & 0.091 & 0.104 & \textbf{0.205} & 0.193 & 0.307 & 0.236 & 0.185 & 0.098 \\
 & Liberal & \textbf{0.148} & 0.092 & \textbf{0.169} & \textbf{0.124} & \textbf{0.142} & 0.181 & 0.165 & 0.273 & \textbf{0.278} & 0.261 & 0.207 \\
 & Conservative & 0.135 & \textbf{0.106} & 0.15 & 0.116 & 0.129 & 0.181 & \textbf{0.239} & 0.295 & 0.274 & 0.163 & 0.163 \\
 & Teenagers & 0.077 & 0.057 & 0.114 & 0.073 & 0.1 & 0.17 & 0.156 & \textbf{0.318} & 0.162 & 0.207 & 0.141 \\
 & People Over 30 & 0.076 & 0.061 & 0.087 & 0.067 & 0.082 & 0.158 & 0.138 & 0.216 & 0.236 & 0.163 & 0.163 \\
    \midrule
    
    \bottomrule
    \end{tabular}}
    \caption{BOLD dataset evaluation. \textbf{Highlighted} values correspond to the highest proportions of prompt completions with a positive/negative sentiment or regard. Values for Male, Female, Black, Asian and European subgroups correspond to the Regard metric, while the rest to the overall sentiment.}
    \label{tab:bold_results_upd}
    \vspace{-4.5mm}
\end{table*}

\subsection{Measuring Bias}

As indicated in the previous section, our development process was mostly guided by qualitative evaluations to chose between alternative approaches and model variants. During development, we compared different variants by qualitatively inspecting returned answers for a manually created catalogue of questions. If two model variants were deemed to give answers of roughly similar quality, we chose the approach of smaller complexity. 


\subsubsection{Qualitative Evaluation}

Table~\ref{tab:answers_short} gives a shortened overview of model outputs for 5 questions and all 11 biases. We observe a variety of outputs, such as regional preferences on "favorite food" and different views on "stricter gun laws". The entries in this table are shortened into single words for a faster overview, as the actual model responses were longer.

Expanding on the question regarding "stricter gun laws", Table~\ref{tab:gunlaws_table} shows the full model responses. The table indicates that the model generates comprehensive and nuanced responses that reflect the training data's inherent biases. In some instances, it constructs responses based on underlying political ideology, demonstrating its understanding of the connection between individual biases and broader political contexts. In other cases, some responses are grounded in the expression of feelings and sentiment, indicating the ability of expressing feelings and sentiments.


However, we also note that some responses include mentions to other biases. For instance, the "Women" answer in Table~\ref{tab:gunlaws_table} is written mainly from a standpoint of a person from Australia. This gives indication to several potential limitations of our approach: First, people posting in a specific subreddit will likely not accurately represent the full demographic we hope to cover (meaning we only model the subset of each demographic that actually posts on Reddit). 
Second, the multifaceted nature of biases - encompassing geographical, political, and age-related diversity among female Reddit users - introduces multiple layers of bias overlap. Resulting in a conflated training signal that potentially leads to less clear bias boundaries in our tuned model. 






\subsubsection{Quantitative Evaluation}

We also experimented with quantitative evaluations to better understand whether each bias group in our model inherently carries a certain view in various political and societal issues, as well as attitude towards different demographics. In order to quantify these notions, we relied on the BOLD dataset~\cite{bold}. It consists of Wikipedia prompts corresponding to different races, genders, religious beliefs, political ideologies, and professions. We use the "regard" metric~\cite{regard} to quantify the attitude of each modeled bias group towards a certain demographic, and regular sentiment analysis~\cite{camacho-collados-etal-2022-tweetnlp} for prompt completions related to political ideologies or religious beliefs.  


Table \ref{tab:bold_results_upd} lists the results for a subset of the BOLD dataset. Overall we observe that the "Liberal" bias exhibits the highest share of prompt completions with negative regard towards four out of five race and gender demographics in the BOLD dataset. Somewhat surprisingly, it also has the second-highest share of prompt completions with negative sentiment towards Liberalism. 

Meanwhile, modeled biases related to women and mature demographics (People Over 30) tend to have a more positive sentiment and regard towards the subgroups and ideas considered in Table \ref{tab:bold_results_upd}. This may reflect usage of a more polite language by these demographics on Reddit. 

\section{Web Demonstration}
\label{sec:demo}

The web-based user interface for OpinionGPT provides an interactive platform for users to interact with the model. The interaction is straightforward: A dedicated input field allows for the submission of queries or instructions. Additionally, users choose from 4 bias groups representing 11 biases supported by the model by clicking the respective selection group (see Figure~\ref{fig:example-output}, upper half). The model then outputs responses for each bias in a selected bias group to the entered question (see Figure~\ref{fig:example-output}, lower half). Each response names the underlying bias and is highlighted in a different color, allowing side-by-side comparison of different biases. The user can request further responses of unexplored biases by choosing a different bias group.

Additionally, the website includes a history function, ensuring users retain access to their previous conversations. The history can be referred to at any time. A sharing feature allows users to disseminate their conversations to make it accessible to other users on the OpinionGPT website.


We build the website upon the open-source project \textit{Chat UI}\footnote{ChatUI: \url{https://github.com/huggingface/chat-ui}} by HuggingFace, albeit heavily modified and customized to align with the unique
needs of OpinionGPT. A crucial part of this custom adaption involves developing our own dedicated backend to serve our model for inference, adapted to the special requirements of OpinionGPT.

\section{Related Work}

\noindent
\textbf{Assessment and Measurement of Biases}.
Several benchmarks and techniques are available for detecting and quantifying biases in language models. StereoSet \citep{nadeem-etal-2021-stereoset} serves as a benchmark to gauge stereotypical bias by assessing language model responses to sentences tied to various demographic groups and stereotypes. The Semantic-associative Evaluation Toolkit (SEAT) \citep{kaneko2021debiasing} quantifies bias by examining the strength of association between pairs of words and attributes. CrowS-Pairs \citep{nangia2020crowspairs} discerns societal biases by evaluating the model's capacity to detect biased sentences within given pairs. 

\noindent
\textbf{Techniques for De-biasing Language Models.}
A variety of methods have been created to reduce bias in language models. For instance, \citep{selfdebias} utilizes self-knowledge distillation to implicitly discern multi-view feature sets, aiming to minimize language bias. SentenceDebias \citep{liang-etal-2020-towards} targets social biases at the sentence-level representation by contextually processing bias-attribute words through a diverse array of sentence templates. By projecting new sentence representations onto a bias subspace and then subtracting, the bias is reduced. More recently, FineDeb \citep{saravanan2023finedeb} was introduced, which employs task-specific fine-tuning on a model pre-trained on extensive text corpora. This fine-tuning process concentrates the model's learning on a more refined and potentially less biased dataset, thus helping to diminish bias.

\noindent
\textbf{Human Alignment for Bias Mitigation.}
Alignment approaches are also utilized to mitigate biases in LLMs. While Instruction Fine-tuning \cite{wei2022flan} trains the model to generate text sequences in a specific format, human alignment, on the other hand, incorporates direct human feedback to shape the model's behavior, utilizing optimization techniques like PPO \cite{ppo} or DPO \cite{dpo}. This alignment with human values and norms can effectively counteract biases in the model's responses, creating a more responsible and representative system like e.g. in the case of Chat-Llama-2 \cite{llama2}.

\section{Conclusion and Discussion}
\vspace{-1mm}

In this paper, we presented OpinionGPT, a web demonstration that allows users to interact with an LLM that was trained on text of different biases. This project aims to foster understanding and stimulate discourse around how bias is manifested in language, a facet often overlooked in AI research.

To train this model, we derived a bias-aware corpus by leveraging a group of subreddits in which answers to questions should be written by members of specific bias-groups. Using this corpus, we fine-tuned a LLaMa model using a designated prompt. This allows us to request answers from the model for specific biases. Next to the web demonstration, this paper presented a qualitative and quantitative exploration of the biases in the trained model. 


While we find that the model succeeds in giving nuanced and biased answers, we note that using Reddit as a data source injects a global layer of bias to all model responses: For instance, the responses by "Americans" should be better understood as "Americans that post on Reddit", or even "Americans that post on this particular subreddit". Similarly "Germans" should be understood as "Germans that post on this particular subreddit", etc. Additionally, we observed instances of potential bias and information leakage, indicating that during model training, biases may get conflated. Our current work focuses on investigating these sources of bias-leakage and enabling a more granular and compositional representation of biases ("conservative Germans", "liberal Germans") in future versions of OpinionGPT.



\section*{Ethics Statement}

As developers of OpinionGPT, we understand and acknowledge the ethical implications that emerge from our work. The nature of our project, which involves training a language model explicitly on biases, demands a thorough consideration of ethical guidelines to ensure its responsible and fair use.

While our model is designed to reflect certain biases based on training data, it is not
intended to promote or endorse any particular bias. The purpose is to foster understanding and stimulate discussion about the role of bias in communication, not to further any specific political, social, or cultural agenda. Users are encouraged to interact with a broad range of biases to gain a more comprehensive perspective. 

We are also mindful of the potential for misuse of our models. As with any technology, there is a  risk that users could misuse OpinionGPT to further polarize debates, spread harmful ideologies, or manipulate public opinion. We therefore made the decision not to publicly release our model. Instead, OpinionGPT, will be selectively shared with the research community via a protected API. 

We are committed to data privacy and protection. Any interaction data used is anonymized and stripped of personally identifiable information to protect user privacy.


\bibliography{custom}

\appendix

\section{Appendix}
\label{appendix1}

\subsection{Instruction Prompt}
The prompt predominantly contains the subreddit name. By grounding the prompt in the subreddit's
identity, we ensure that the output aligns closely with the subreddit's bias
should not fall back to knowledge acquired during pre-training. Our chosen
prompt:

\begin{lstlisting}
### r/{subreddit} Question:

{instruction}

### r/{subreddit} Answer:


\end{lstlisting}

\subsection{Screencast Video}
A screencast video demonstrating OpinionGPT is available under:
\url{https://vimeo.com/886419062}

\label{sec:appendix}

\end{document}